\RequirePackage[svgnames]{xcolor}

\documentclass[11pt,a4paper]{mystyle}

\usepackage[all]{hypcap}
\usepackage[svgnames]{xcolor}
\usepackage[comma,authoryear,compress]{natbib}
\renewcommand{\cite}{\citep}

\usepackage{arydshln}
\usepackage{algpseudocode}
\usepackage{microtype}
\usepackage{booktabs}
\usepackage{float}
\usepackage{bigstrut}
\usepackage{multirow}
\usepackage{amsmath}
\usepackage{amssymb}
\usepackage{mathtools}
\usepackage{amsthm}
\usepackage{mathrsfs}
\usepackage{dsfont}
\usepackage{enumitem}
\usepackage{graphicx}
\usepackage{cleveref}
\usepackage{bxcoloremoji}

\usepackage{float}

\usepackage[utf8]{inputenc} %
\usepackage[T1]{fontenc}    %
\usepackage{hyperref}       %
\usepackage{url}            %
\usepackage{amsfonts}       %
\usepackage{nicefrac}       %
\usepackage{microtype}      %
\usepackage{fdsymbol}
\usepackage{wrapfig}
\usepackage{lipsum}
\usepackage{stackengine}
\usepackage[font=small,labelfont=bf]{caption}
\usepackage{color}
\usepackage{adjustbox}

\usepackage{rotating}
\usepackage{makecell}

\title{Enhancing Fake News Video Detection via LLM-Driven Creative Process Simulation}

\runningtitle{Enhancing Fake News Video Detection via LLM-Driven Creative Process Simulation}

\author[1,2]{\href{https://scholar.google.com/citations?user=i9SLGsEAAAAJ/}{\textcolor{black}{Yuyan Bu}}}
\author[1]{\href{https://sheng-qiang.github.io/}{\textcolor{black}{Qiang Sheng}}}
\author[1,2]{\href{https://scholar.google.com/citations?user=fSBdNg0AAAAJ/}{\textcolor{black}{Juan Cao}}}
\author[1,2]{\href{https://scholar.google.com/citations?&user=clCHEnkAAAAJ/}{\textcolor{black}{Shaofei Wang}}}
\author[3]{\href{https://scholar.google.com/citations?&user=cJ85jdEAAAAJ/}{\textcolor{black}{Peng Qi}}}
\author[1,2]{\href{https://scholar.google.com/citations?&user=xAw_fukAAAAJ/}{\textcolor{black}{Yuhui Shi}}}
\author[1,2]{\href{https://beanandrew.github.io/}{\textcolor{black}{Beizhe Hu}}}

\affil[1]{Media Synthesis and Forensics Lab, Institute of Computing Technology, Chinese Academy of Sciences}
\affil[2]{University of Chinese Academy of Sciences}
\affil[3]{National University of Singapore}

\emails{{\href{mailto:jzdbyy@gmail.com}{jzdbyy@gmail.com},  \{\href{mailto:shengqiang18z@ict.ac.cn}{shengqiang18z},\href{mailto:caojuan@ict.ac.cn}{caojuan},\href{mailto:wangshaofei24s@ict.ac.cn}{wangshaofei24s}\}@ict.ac.cn, \href{mailto:peng.qi@nus.edu.sg}{peng.qi@nus.edu.sg}, \{\href{mailto:shiyuhui22s@ict.ac.cn}{shiyuhui22s},\href{mailto:hubeizhe21s@ict.ac.cn}{hubeizhe21s}\}@ict.ac.cn}}

\begin{document}

\begin{abstract}
  The emergence of fake news on short video platforms has become a new significant societal concern, necessitating automatic video-news-specific detection. 
  Current detectors primarily rely on pattern-based features to separate fake news videos from real ones. However, limited and less diversified training data lead to biased patterns and hinder their performance. 
  This weakness stems from the complex many-to-many relationships between video material segments and fabricated news events in real-world scenarios: a single video clip can be utilized in multiple ways to create different fake narratives, while a single fabricated event often combines multiple distinct video segments. 
  However, existing datasets do not adequately reflect such relationships due to the difficulty of collecting and annotating large-scale real-world data, resulting in sparse coverage and non-comprehensive learning of the characteristics of potential fake news video creation.
  To address this issue, we propose a data augmentation framework \textbf{AgentAug} that generates diverse fake news videos by simulating typical creative processes. AgentAug implements multiple LLM-driven pipelines of four fabrication categories for news video creation, combined with an active learning strategy based on uncertainty sampling to select the potentially useful augmented samples during training.
  Experimental results on two benchmark datasets demonstrate that AgentAug consistently improves the performance of short video fake news detectors.
\vspace{5mm}

\coloremojicode{1F4C5} \textbf{Date}: October 5, 2025

\coloremojicode{1F3E0} \textbf{Project}: \href{https://github.com/ICTMCG/AgentAug}{https://github.com/ICTMCG/AgentAug}

\coloremojicode{1F4AC} \textbf{Venue}: ACM CIKM 2025 (\href{https://doi.org/10.1145/3746252.3760933}{https://doi.org/10.1145/3746252.3760933})

\end{abstract}

\maketitle
\vspace{3mm}

\section{Introduction}

Though short video platforms facilitate users to get news more conveniently~\cite {hendrickx2023newspapers,niu2023building,TikTokNewsConsumption}, their characteristics of rapid, recommendation-driven content dissemination also amplify misinformation risks~\cite{bu2023combating}. Therefore, fake news detection requires an urgent transformation from text or image-based~\cite{zoomout,genfend,qipeng-chapter,ma-etal-2024-event} to addressing video challenges.
Existing methods in fake news video detection mainly rely on mining discriminative patterns reflected by training data~\cite{papadopoulou2017,li2022cnn,palod2019vavd,shang2021tiktec}, which is effective under ideal conditions but vulnerable due to the inherent dependence on data quality and coverage diversity.  When the training dataset is small and fails to reflect the complexity of real-world scenarios, the learned detector will hardly identify the characteristics of fake news videos in diverse forms. 

\begin{figure}[t]  
    \centering
    \includegraphics[width=0.9\linewidth]{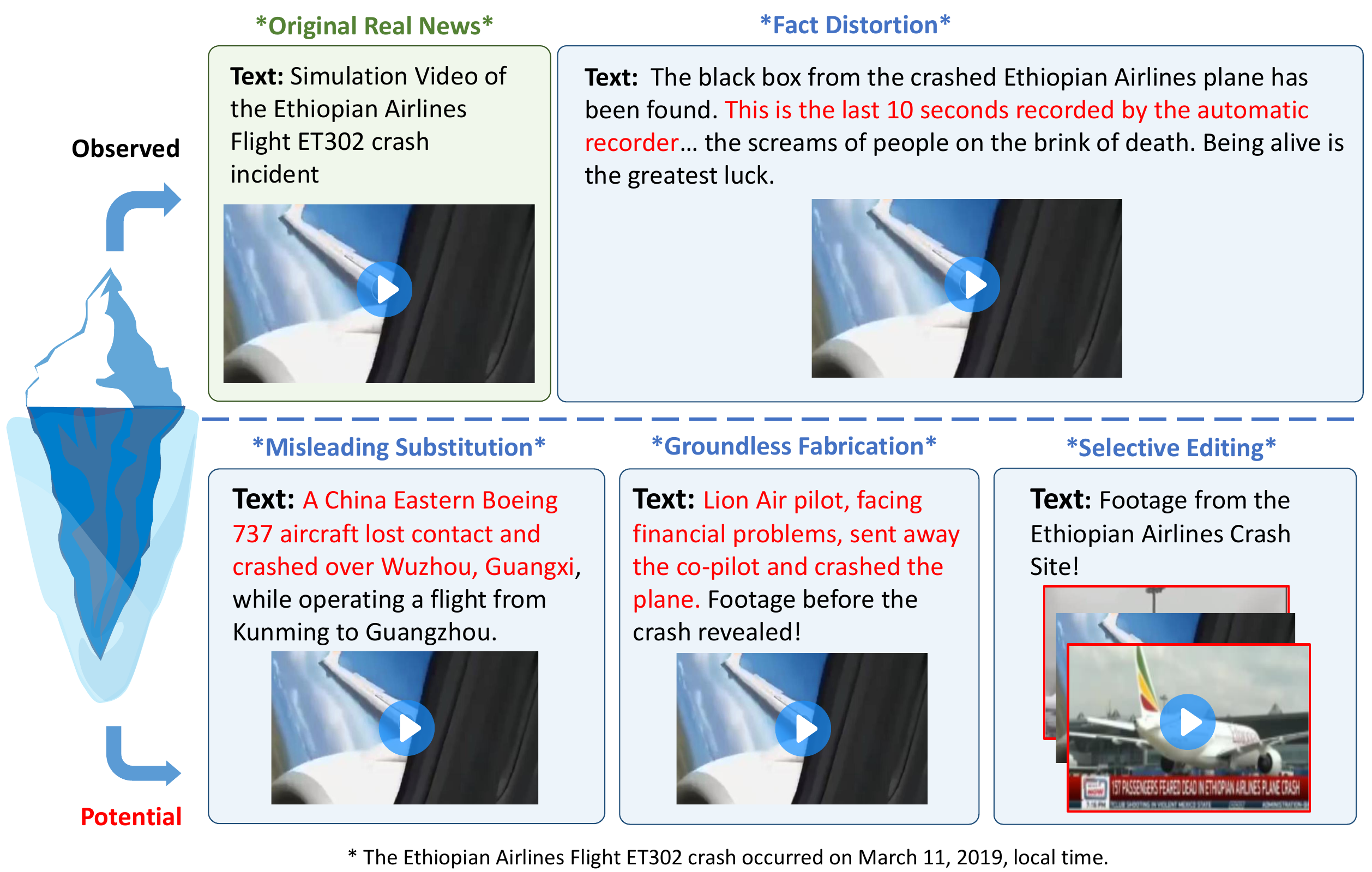}
    \caption{An example of how original real materials can be used for fabrication in multiple ways (highlighted in red), while some may not be shown in existing observed data.}
    \label{fig:headfig} 
\end{figure}

Such complexity is rooted in the creative process of short video fake news. The possible fabrication ideas of fake news often shape a \textit{many-to-many} relationship between the source materials and the event depicted.
As exemplified by Figure~\ref{fig:headfig}, a single source material of simulated footage of the Ethiopian Airlines crash~\cite{simulation} can be manipulated through diverse fabrication techniques to spawn multiple fake narratives: using Fact Distortion, the footage could be misrepresented as real footage of a black-box recording; using Selective Editing, unrelated news clips could be inserted into the video to alter context; using Groundless Fabrication, a fictional narrative “Lion Air pilot deliberately crashed over financial disputes” is added; and with Misleading Substitution, the footage is repurposed to report another unrelated event “China Eastern Airlines crash.”
Conversely, a single news event can also be constructed from multiple video clips with various fabrication techniques. Such many-to-many mapping causes high diversity of fake news characteristics, but the coverage of potential manipulation permutations in existing datasets~\cite{papadopoulou2019corpus,qi2022fakesv,fakingrecipe} inevitably remains sparse.
For a fake event in reality, only some of the fabrications are applied and observed.
Such data insufficiency induces models to learn biased feature correlations specific to particular material-event co-occurrences rather than generalizable detection principles.
When confronted with other fabrications, detectors trained on such real-data distributions may suffer performance degradation due to their over-reliance on spurious patterns~\cite{wan2025truth}.

To overcome this limitation, increasing the scale and diversity of training data presents an intuitive solution. However, the time and financial costs of large-scale real-world annotated data acquisition make this way almost infeasible.
In this paper, we propose to automatically synthesize diverse fake news samples to simulate the many-to-many mappings between source materials and fabricated events, thereby alleviating performance bottlenecks caused by insufficient diversity in training data, resulting in the \textbf{AgentAug} framework shown in Figure~\ref{fig:agentaug_overall_pipeline}.

Before constructing AgentAug, we performed a systematic analysis of common human fabrication techniques, from which we distill two fundamental categories of fake news generation (completion-based and fabrication-based) along with four basic fabrication actions: fact distortion, groundless fabrication, misleading substitution, and selective editing, which has been exemplified in Figure~\ref{fig:headfig}.
Building on this, AgentAug builds an automated synthesis pipeline that uses these four core fabrication paradigms to generate diverse fake news samples based on existing materials, as a supplement for the samples typically not covered before. We first construct a material library based on existing datasets, which provides materials for the subsequent pre-designed synthesis pipelines to produce various fake news samples. The pipelines are powered by large language models~(LLMs)~\cite{tan2024large} to ensure the diversity and rationality of the produced samples.
In data use, we introduce an active learning-based sampling strategy to adaptively select suitable samples from the data pool based on the specific requirements of trained models. Experiments conducted on two real-world datasets demonstrate the effectiveness of AgentAug in improving detection performance. Our main contributions are as follows:
\begin{itemize}
    \item \textbf{Idea:} We propose to generate data to alleviate the insufficiency issue in fake news video detection by building multiple crafted agentic workflows as an enhancement.
    \item \textbf{Method:} We build \textbf{AgentAug}, consisting of four LLM-driven agentic workflows for data synthesis and an active learning strategy for data selection, which is the first in this area.
    \item \textbf{Performance:} Experiments on two real-world datasets show that AgentAug can improve the detection performance of four representative detectors consistently.
\end{itemize}

\section{Methodology}
\begin{figure}[t]  
    \centering    
\includegraphics[width=0.9\linewidth]{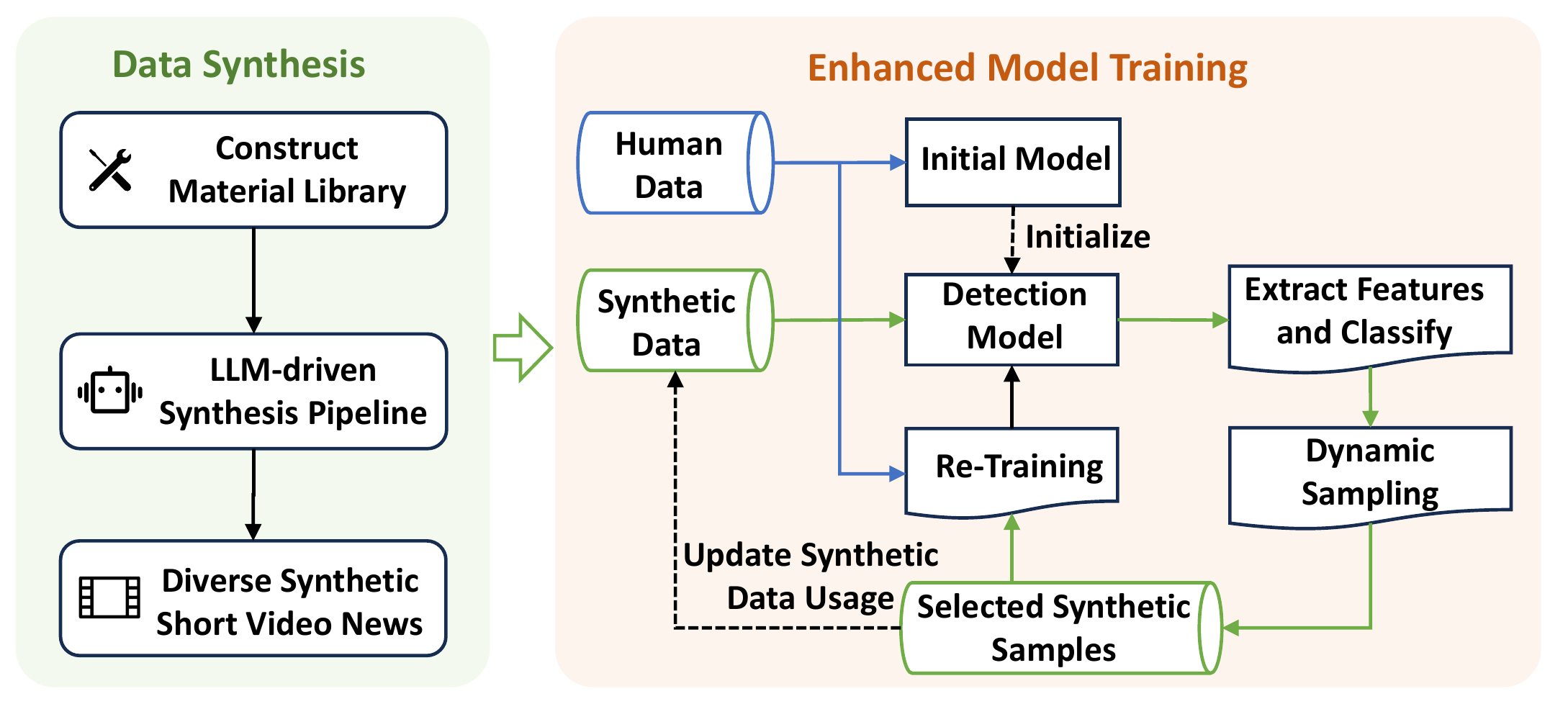}
    \caption{Overall framework of AgentAug.}\label{fig:agentaug_overall_pipeline} 
\end{figure}

As shown in Figure~\ref{fig:agentaug_overall_pipeline}, the overall framework comprises two main parts: Data Synthesis and Enhanced Model Training. We first construct a material library and develop an LLM-driven pipeline to establish an agentic workflow, enabling efficient data generation. Then, the model training adopts an active learning strategy to dynamically sample the generated data for performance optimization.

\subsection{Agentic Workflows for Data Synthesis}
\label{sec:data-synthesis}

\noindent\textbf{Material Library Construction.}
Given that video news primarily conveys information through text and visual content, the material library is constructed by collecting textual and visual materials. 

For the textual material library, we employ Qwen-Max~\cite{qwen1.5} to analyze captions of real-world news short videos by filtering out obscure texts with little informational value for video event presentation. Based on the training sets of the benchmark datasets FakeSV (Chinese) and FakeTT (English), we archive 2,106 (Chinese) and 910 (English) news text samples, respectively.

\begin{figure}[t]  
    \centering    
\includegraphics[width=0.8\linewidth]{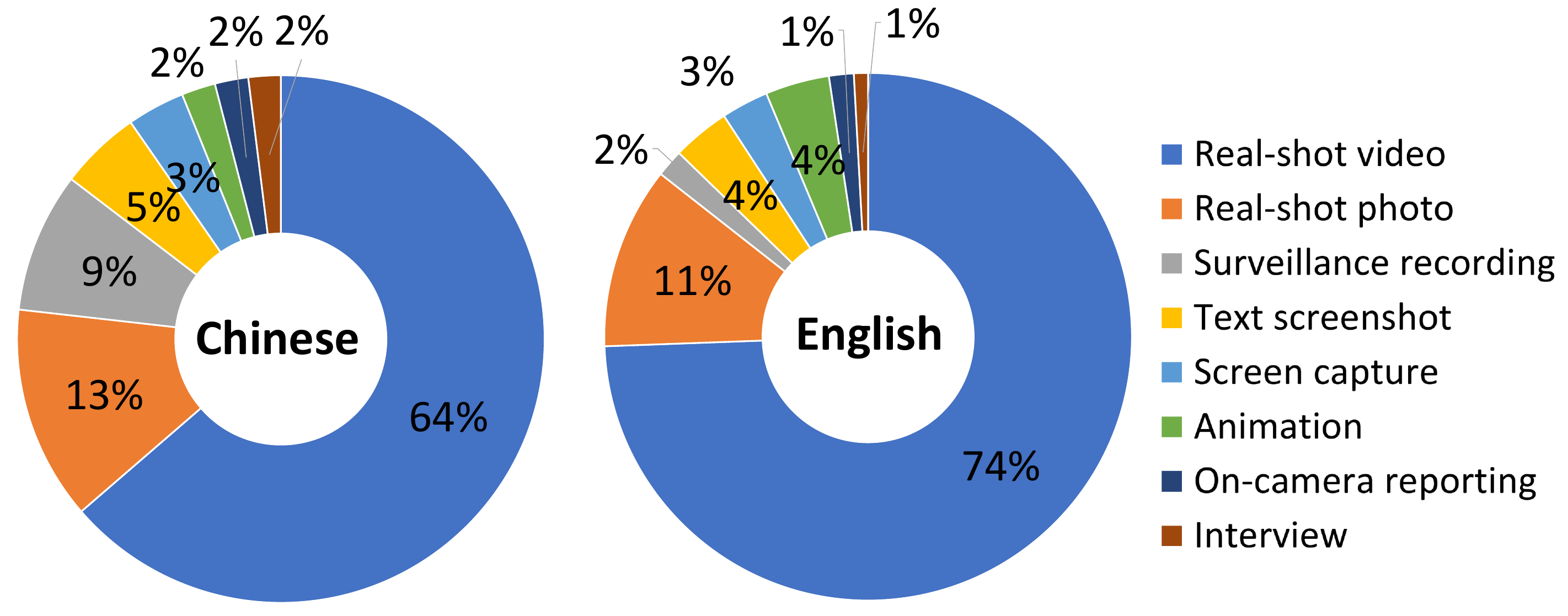}
    \caption{Distribution of visual material types in the constructed video clip libraries.}
    \label{fig:vmaterial_type} 
\end{figure}

For the visual material library, we first utilize the shot segmentation model TransNet-v2~\cite{transnetv2} to segment existing news videos into raw visual clips. To facilitate subsequent material selection, these clips are further processed using Qwen-VL-Max~\cite{Qwen-VL} for content-aware understanding and archival. Specifically, Qwen-VL-Max generates objective descriptions of the clips based on key visual elements (e.g., ``oil tank'', ``flames'', ``water spray'' in a fire scene) and categorizes them into types (e.g., ``real-shot video'', ``real-shot photo'', ``interview'', ``screenshot'', etc.).
Additionally, it evaluates each clip across visual quality, newsworthiness, and visual impact by assigning scores from 1 to 5 (higher is better) with rational justifications. 
Figure~\ref{fig:vmaterial_type} illustrates the distribution of material types in the video clip libraries. 
After filtering out short, trivial, or meaningless clips, we archive 11,095/6,294 video news clips from the training set of FakeSV/FakeTT, forming a solid foundation for the subsequent synthesis process.

\begin{figure}[t]  
    \centering    
\includegraphics[width=1\linewidth]{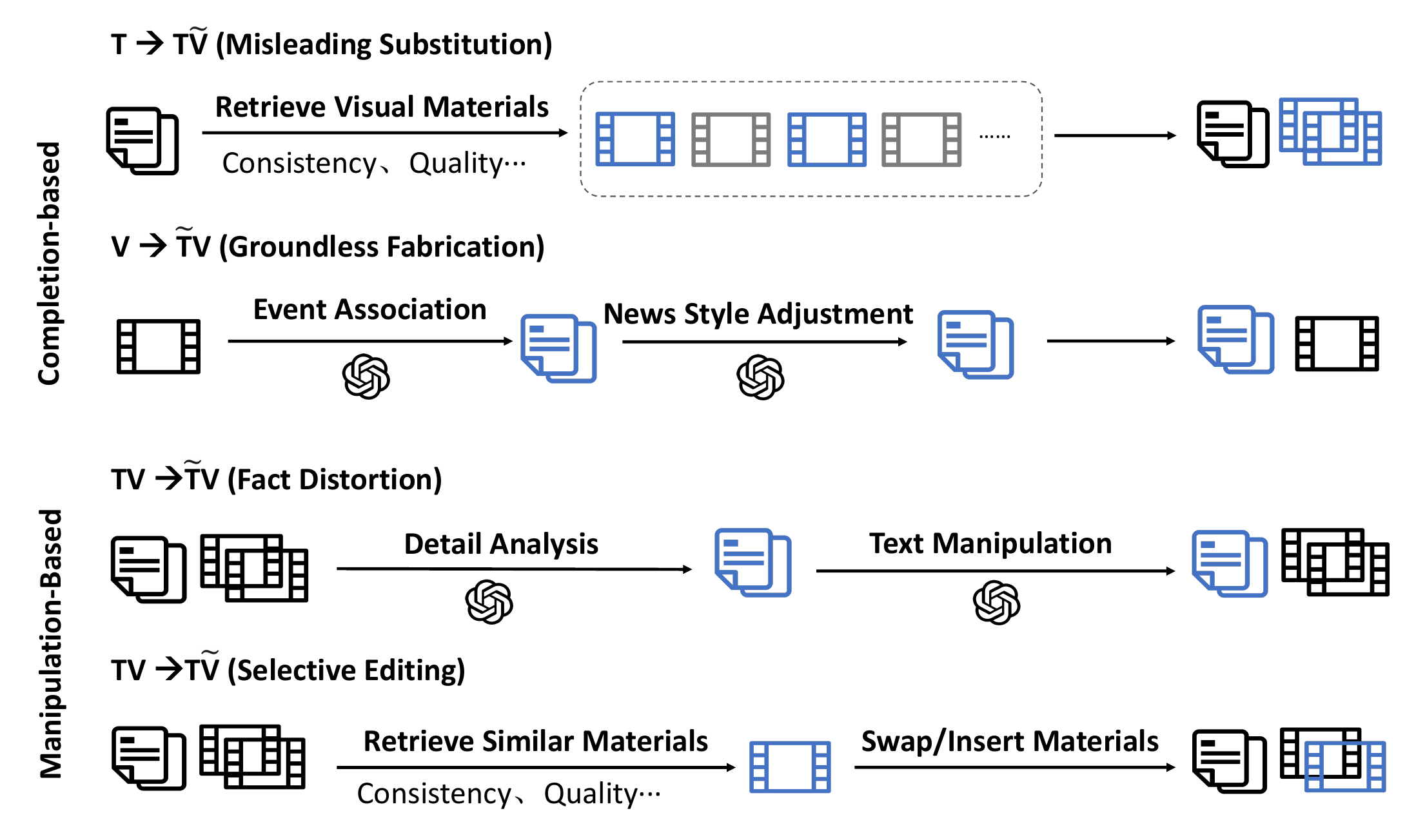}    \caption{LLM-driven pipelines for video news fabrication.}
    \label{fig:agentmade_pipeline} 
\end{figure}

\begin{figure*}[t]  
    \centering    
\includegraphics[width=1\linewidth]{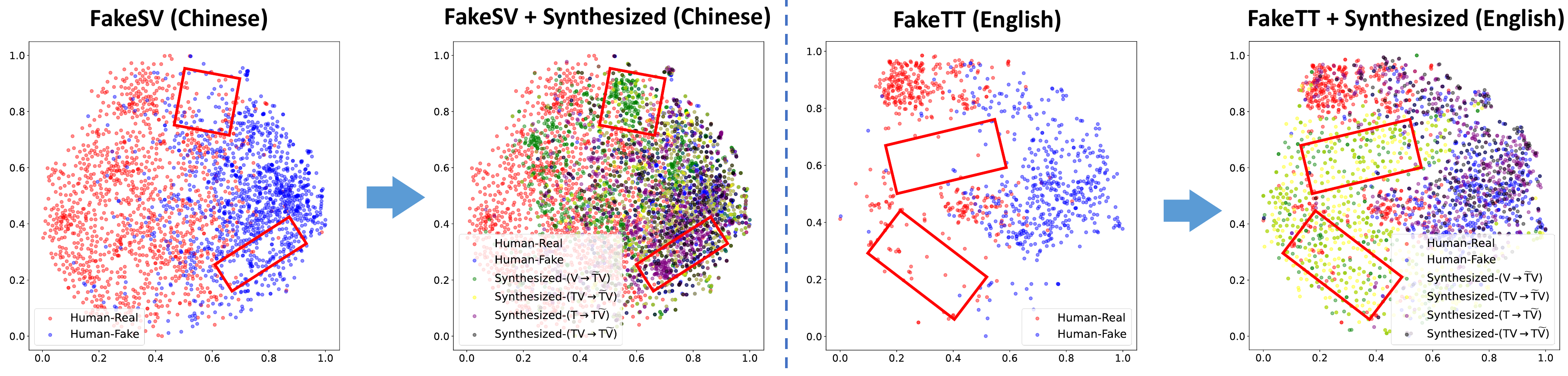}
    \caption{T-SNE visualization of data samples in the feature spaces. Red boxes highlight the effects of synthesized data.}
    \label{fig:fea_map} 
\end{figure*}

\noindent\textbf{LLM-driven Pipeline.}
To design an effective data synthesis pipeline, we first analyze typical human fabrication strategies.
Fake news creation is often triggered by specific source materials, which may consist of unimodal text (T), unimodal visual content (V), or multimodal content (TV). 
Note that we focus on textual and visual content only, excluding audio~\cite{liu2024exploring} and metadata~\cite{qi2022fakesv}, as the former two modalities have expressed rich semantics.
Fabricated outputs are usually obtained by modifying or supplementing one or more modalities of the original materials, resulting in news that diverges from or contradicts the original event. Such manipulations include altering only the text ($\widetilde{\mathrm{T}}$V) or visual content (T$\widetilde{\mathrm{V}}$), and distorting both modalities ($\widetilde{\mathrm{T}}$$\widetilde{\mathrm{V}}$). 
We focus on modeling four representative fabrication logics that align with common human practices in fake video creation: V$\rightarrow$$\widetilde{\mathrm{T}}$V and T$\rightarrow$T$\widetilde{\mathrm{V}}$ (completion-based), and TV$\rightarrow$$\widetilde{\mathrm{T}}$V and TV$\rightarrow$T$\widetilde{\mathrm{V}}$ (manipulation-based). These cover the core patterns in fake news creation, including 
groundless fabrication, misleading substitution, fact distortion, and selective editing.

The LLM-driven pipeline designed according to these four strategies is illustrated in Figure~\ref{fig:agentmade_pipeline}.
For T$\rightarrow$T$\widetilde{\mathrm{V}}$, we retrieve candidate visual clips by matching text and visual descriptions based on the BGE embedding~\cite{bge_m3} and iteratively select and append relevant ones until the target duration is reached.
For V$\rightarrow$$\widetilde{\mathrm{T}}$V, we exploit LLMs' capacity to generate fictional news events from visual input, producing news-style coherent text descriptions.
For TV$\rightarrow$T$\widetilde{\mathrm{V}}$, we replace or insert visual clips to introduce misleading visual evidence by first scoring original video segments using the metrics mentioned in Section~\ref{sec:data-synthesis} and replacing low-scored ones with high-scored, non-overlapping segments in the library.
For TV$\rightarrow$$\widetilde{\mathrm{T}}$V, we prompt the LLM to modify factual details while preserving semantic consistency with the visual content.
Applied to benchmark datasets, the pipeline synthesizes 3,278 augmented fake news videos based on FakeSV and 2,592 based on FakeTT.

\noindent\textbf{Analysis of Synthesized Data.}
To assess the effectiveness of our synthesized data, we adopt t-SNE~\cite{tsne} to visualize the feature distributions of synthesized and human data in the SVFEND~\cite{qi2022fakesv} feature space, as displayed in Figure~\ref{fig:fea_map}.

We observe that human and synthesized data points show a high degree of overlap overall. Such a close alignment in feature space indicates that the synthesized data effectively mirrors the distribution of human-produced samples and underscores its validity.
Moreover, as highlighted by the red-boxed regions, the synthesized samples not only overlap with human ones but also populate sparse areas of the feature space, thus improving the space coverage. This expansion demonstrates the potential value of the augmented data in making detectors learn more diverse fabrication types and improving detection performance.

\subsection{Enhanced Training through Active Learning}

Not all synthesized data needs to be included for training because of the differences in the models' capacity and weaknesses.
Therefore, we design an active learning-based retraining framework that incrementally selects and incorporates informative synthesized samples based on model uncertainty and feature similarity.

The procedure begins with a detector trained on human-annotated data. To ensure robustness and mitigate overfitting to a fixed validation set, we adopt a five-fold cross-validation strategy, using four folds for training and the remaining one for validation in each round. For each iteration within a fold, the model estimates prediction uncertainty on the validation set using entropy: $H(p)=-plogp-(1-p)log(1-p)$, where $p$ is the predicted probability of a sample being fake. Higher entropy indicates that the model is less confident in its prediction, and thus the sample is more informative for active selection~\cite{tang-etal-2002-active}. The top $\alpha$ fraction of samples with the highest entropy are selected as anchors. 
We then retrieve the top-$K$ similar synthesized samples from a candidate pool, based on cosine similarity in the feature space extracted by the current model. These samples are temporarily added to the training set for one iteration of retraining. If the updated model improves validation performance (measured by F1 on anchor samples), these synthesized instances are retained; otherwise, they are discarded or partially rejected.

This process is repeated for $T$ iterations within each fold. After all five folds are completed, we aggregate selection frequencies across folds and retain only those synthesized samples selected in at least three folds. The final model is trained on the union of the original labeled data and the consistently selected synthesized samples. Note that this procedure is compatible with various detection backbones, only requiring that the detector is learnable and makes predictions with probabilities.
More detailed illustrations, analysis, and framework implementation are in the project repository.

\section{Experiments}
In this section, we conduct extensive experiments on two real-world datasets to evaluate the effectiveness of the proposed data augmentation method. The method is applied across four representative baseline detectors to assess its generalizability.

\subsection{Experimental Setup}

\begin{table}[t]
\centering
\caption{Statistics of two experimental datasets.}
\label{dataset_statistic}
\begin{tabular}{cccrrr} 
\toprule
Dataset & Time Range & Avg Duration (s) & \#Fake & \#Real & \#All  \\ 
\midrule
FakeSV  & 2017/10-2022/02 & 39.88       & 1,810   & 1,814   & 3,624   \\
FakeTT  & 2019/05-2024/03 & 47.69       & 1,172   & 819    & 1,991   \\
\bottomrule
\end{tabular}
\end{table}

\noindent\textbf{Datasets.}
We conduct experiments on two public real-world datasets, FakeSV~\cite{qi2022fakesv} and FakeTT~\cite{fakingrecipe} (Table~\ref{dataset_statistic}):

\textit{FakeSV}\footnote{\url{https://github.com/ICTMCG/FakeSV}} is a publicly available Chinese dataset for fake news detection on short video platforms. It contains videos collected from \textit{Douyin} and \textit{Kuaishou}, two widely used video-sharing platforms in China. Each sample includes the video, its title, user comments, metadata, and publisher profiles.

\textit{FakeTT}\footnote{\url{https://github.com/ICTMCG/FakingRecipe}} is an English dataset curated from \textit{TikTok}. It follows a similar construction process to FakeSV and provides each video sample along with its audio and text description.

Following~\citet{qi2022fakesv}, we adopt a temporal split (train:val:test=70\%/15\%/15\%) to simulate realistic deployment settings. For fair evaluation, the test set only contains human-produced videos, and all synthesized samples are strictly generated based on training data to avoid potential leakage.

\noindent\textbf{Baselines and implementation details.} 
The proposed AgentAug method is applied on four representative baseline detectors to assess its effectiveness: FANVM~\cite{choi2021cikm}, SVFEND~\cite{qi2022fakesv}, SVRPM~\cite{wu2024interpretable}, and MMVD~\cite{zeng2024mitigating}. When implementing baselines, we remove the social context part (if any) for a fair comparison. We did a search in [100, 1000] with a step of 100 to set a proper $K$ based on validation performance. $T$ is the minimum number to ensure all synthesized data is considered at once under the selected $K$. The duration is set to 15 seconds. $\alpha$ is set to 30\%.

\subsection{Results and Analysis}

\noindent\textbf{Main Results.}
Table~\ref{tab:AgentAug_ablation_bi} reports the detection performance before and after applying AgentAug (with subscript AL). Overall, the augmented data generated by our automatic pipeline consistently improves model performance across all baselines. We observe that weaker detectors (e.g., FANVM) benefit more significantly, likely because their limited learning capacity requires more straightforward signals from data, which can be provided by our generations with diverse narrative strategies.
In contrast, stronger models (e.g., SVRPM) exhibit more moderate yet stable improvements, indicating that AgentAug still helps refine decision boundaries.

\begin{table}[t]
\centering
\small
\caption{Detection performance (\%$\uparrow$) on \textbf{FakeSV} (Chinese) and \textbf{FakeTT} (English) with different sampling strategies.}
\label{tab:AgentAug_ablation_bi}
\begin{tabular}{l cccc cccc}
\toprule
\multirow{2}{*}{\textbf{Model}} &
\multicolumn{4}{c}{\textbf{FakeSV}} &
\multicolumn{4}{c}{\textbf{FakeTT}} \\ 
\cmidrule(lr){2-5}\cmidrule(lr){6-9} 
& Acc & F1 & Prec. & Rec.
& Acc & F1 & Prec. & Rec. \\ 
\midrule
\textbf{MMVD}                        & 75.83 & 75.33 & 75.51 & 75.21 & 67.50 & 66.20 & 66.51 & 68.43  \\
w/AgentAug$_{\text{RAN}}$            & 73.80 & 73.22 & 73.33 & 73.13 & 63.57 & 61.69 & 61.84 & 63.11  \\
w/AgentAug$_{\text{BAL}}$            & 76.01 & 75.58 & 75.86 & 75.45 & 66.43 & 65.32 & 65.55 & 66.94  \\
\rowcolor{gray!10}
w/AgentAug$_{\text{AL}}$             & \textbf{77.12} & \textbf{76.69} & \textbf{76.93} & \textbf{76.55} & \textbf{68.57} & \textbf{67.26} & \textbf{67.56} & \textbf{69.71} \\
\cdashline{1-9}[2pt/2pt]
\textbf{FANVM}                       & 78.41 & 77.89 & 78.25 & 77.70 & 71.57 & 70.21 & 70.21 & 72.63 \\
w/AgentAug$_{\text{RAN}}$            & 78.78 & 77.63 & 80.08 & 77.16 & 67.22 & 66.77 & 69.18 & 71.42 \\
w/AgentAug$_{\text{BAL}}$            & 79.34 & 78.31 & 80.42 & 77.84 & 68.23 & 67.99 & 71.57 & 73.70 \\
\rowcolor{gray!10}
w/AgentAug$_{\text{AL}}$             & \textbf{81.37} & \textbf{80.42} & \textbf{82.73} & \textbf{79.88} & \textbf{75.25} & \textbf{74.02} & \textbf{73.84} & \textbf{76.65} \\
\cdashline{1-9}[2pt/2pt]
\textbf{SVFEND}                      & 80.88 & 80.54 & 80.83 & 80.51 & 77.14 & 75.63 & 75.12 & 77.56 \\
w/AgentAug$_{\text{RAN}}$            & 82.66 & 81.94 & 83.54 & 81.44 & 79.72 & 77.72 & 77.33 & 78.24 \\
w/AgentAug$_{\text{BAL}}$            & 81.92 & 80.85 & 83.96 & 80.23 & 77.58 & 76.00 & 75.43 & 77.40 \\
\rowcolor{gray!10}
w/AgentAug$_{\text{AL}}$             & \textbf{83.76} & \textbf{82.98} & \textbf{85.20} & \textbf{82.38} & \textbf{80.43} & \textbf{78.61} & \textbf{78.12} & \textbf{79.29} \\
\cdashline{1-9}[2pt/2pt]
\textbf{SVRPM}                       & 81.34 & 81.11 & 81.38 & 80.97 & 81.79 & 79.42 & 79.67 & 79.19  \\
w/AgentAug$_{\text{RAN}}$            & 79.57 & 79.33 & 79.61 & 79.20 & 80.00 & 77.34 & 77.49 & 77.19  \\
w/AgentAug$_{\text{BAL}}$            & 81.73 & 81.58 & 82.06 & 81.49 & 82.14 & 80.28 & \textbf{81.36} & 79.58  \\
\rowcolor{gray!10}
w/AgentAug$_{\text{AL}}$             & \textbf{83.10} & \textbf{82.89} & \textbf{83.16} & \textbf{82.70} & \textbf{82.86} & \textbf{80.57} & 80.75 & \textbf{80.41}  \\
\bottomrule
\end{tabular}%
\end{table}

\noindent\textbf{Ablation Study.}
We compare our active learning (AL) method with two alternatives: random sampling (AgentAug\textsubscript{RAN}) and class-balanced sampling (AgentAug\textsubscript{BAL}). For fair comparison, all settings are kept the same except the sampling mechanism. AgentAug\textsubscript{RAN} randomly selects the same number of samples from the augmented pool, while AgentAug\textsubscript{BAL} ensures equal counts across four manipulation types through within-class random sampling.

As shown in Table~\ref{tab:AgentAug_ablation_bi}, both AgentAug\textsubscript{RAN} and AgentAug\textsubscript{BAL} yield unstable performance. In some cases (e.g., FANVM on FakeTT), model performance even degrades after augmentation. This suggests that unguided sampling may introduce redundant or low-value samples, limiting the effectiveness of augmented training. Although AgentAug\textsubscript{BAL} mitigates class imbalance to some extent, it fails to consider sample difficulty or model uncertainty. In fact, its forced uniformity may amplify low-quality examples, suppressing performance gains.
In contrast, our active learning-based sampling (AgentAug\textsubscript{AL}) adaptively targets high-uncertainty regions, selecting more informative samples. It consistently outperforms the other strategies across most models and avoids performance drops—demonstrating stronger robustness and data efficiency.

\begin{table}[t]
\centering
\small
\caption{Ablation study on fabrication-type combinations based on the SVFEND detector.}
\label{tab:faking_type_ablation}
\begin{tabular}{ccccccccc}
\toprule
\multirow{2}{*}{\textbf{Settings}} &
\multicolumn{4}{c}{\textbf{FakeSV}} &
\multicolumn{4}{c}{\textbf{FakeTT}} \\
\cmidrule(lr){2-5}\cmidrule(lr){6-9}
& Acc & F1 & Prec. & Rec.  & Acc & F1  & Prec. & Rec. \\
\midrule
ALL                      & \textbf{83.76} & \textbf{82.98} & \textbf{85.20} & \textbf{82.38} & \textbf{80.43} & \textbf{78.61} & \textbf{78.12} & \textbf{79.29} \\
w/o V$\rightarrow \widetilde{\mathrm{T}}$V           & 82.10 & 81.10 & 83.95 & 80.49 & 77.94 & 76.52 & 75.95 & 78.18 \\
w/o T$\rightarrow$T$\widetilde{\mathrm{V}}$           & 79.34 & 79.24 & 79.27 & 79.71 & 77.58 & 76.10 & 75.53 & 77.66 \\
w/o TV$\rightarrow\widetilde{\mathrm{T}}$V           & 82.29 & 81.93 & 82.13 & 81.79 & 79.00 & 76.83 & 76.54 & 77.19 \\
w/o TV$\rightarrow$T$\widetilde{\mathrm{V}}$          & 82.47 & 81.82 & 83.11 & 81.36 & 78.29 & 76.58 & 76.00 & 77.68 \\
\bottomrule
\end{tabular}
\end{table}

Further, Table~\ref{tab:faking_type_ablation} presents a manipulation-type ablation based on the SVFEND detector. Removing any one of the four manipulation types results in lower performance compared to the full AgentAug setting, indicating that the diverse fabrication strategies provide complementary benefits.

\section{Conclusion and Discussion}
We proposed \textbf{AgentAug}, a framework consisting of LLM-driven data synthesis and active learning-based data selection, specifically for training enhancement of fake news video detectors. Empowered by LLMs, we build four fabrication pipelines to simulate the diverse creative process of fake news videos with limited existing materials from the training data. The synthesized data is then selected in a dynamic and on-demand way by an active learning strategy for enhancing the training of fake news video detectors. Based on two widely used public datasets and four representative detection methods, we validate the effectiveness of AgentAug in detection improvement.
We expect that AgentAug exhibits the way of how to synthesize complex multi-modal data to assist model training and advocate for more attention to this direction.

\noindent\textbf{Limitations.}
\textbf{1)} Our LLM-driven synthesis explores multiple practical combinations of existing materials and generations to cover diverse faking types but does not include \textit{purely} generated fake news videos as recently-emerged full generation of fake news videos still faces controllability and quality issues and do not form a mature fabrication way~\cite{lei2024comprehensive}. 
\textbf{2)} Though we have enumerated possible situations in modality completion and modification in video news faking, there might be other perspectives for categorization of fake news videos~\cite{media-lessons}.
\textbf{3)} We build several task-specific agentic workflows for data synthesis but do not explore more autonomous agents which might further improve data diversity based on feedback~\cite{anthropic-agents}.
We will perform further exploration to address the mentioned limitations.

\noindent\textbf{Ethical Considerations.} 
All synthesized fake news samples are created exclusively for research purposes, aiming to improve the detectors' performance by diversifying training data.
To mitigate potential misuse, we will not publicly release complete synthesized news videos.
For reproduction aims in academia, we will share necessary materials based on an application review mechanism.

\section*{GenAI Usage Disclosure}
This study strictly follows the guidelines and conventions of AI tool use when preparing this submission. Specifically, generative models were used to simulate fake news narratives for data augmentation, under the authors' careful supervision and control. Additionally, large language models were utilized to improve the clarity and fluency of the manuscript's wording. All research ideas, methodological designs, and analyses were independently conceived and executed by the authors.

\bibliography{main}

\begin{thebibliography}{31}
\providecommand{\natexlab}[1]{#1}

\bibitem[{Anthropic(2024)}]{anthropic-agents}
Anthropic. 2024.
\newblock Building Effective Agents.
\newblock
  \url{https://www.anthropic.com/engineering/building-effective-agents}.

\bibitem[{Bai et~al.(2023)Bai, Bai, Yang, Wang, Tan, Wang, Lin, Zhou, and
  Zhou}]{Qwen-VL}
Jinze Bai, Shuai Bai, Shusheng Yang, Shijie Wang, Sinan Tan, Peng Wang, Junyang
  Lin, Chang Zhou, and Jingren Zhou. 2023.
\newblock \href {https://arxiv.org/abs/2308.12966} {{Qwen-VL: A Versatile
  Vision-Language Model for Understanding, Localization, Text Reading, and
  Beyond}}.
\newblock \emph{Preprint}, arXiv:2308.12966.

\bibitem[{Bu et~al.(2023)Bu, Sheng, Cao, Qi, Wang, and Li}]{bu2023combating}
Yuyan Bu, Qiang Sheng, Juan Cao, Peng Qi, Danding Wang, and Jintao Li. 2023.
\newblock \href {https://doi.org/10.1145/3581783.3612426} {{Combating Online
  Misinformation Videos: Characterization, Detection, and Future Directions}}.
\newblock In \emph{Proceedings of the 31st ACM International Conference on
  Multimedia}, pages 8770--8780. Association for Computing Machinery.

\bibitem[{Bu et~al.(2024)Bu, Sheng, Cao, Qi, Wang, and Li}]{fakingrecipe}
Yuyan Bu, Qiang Sheng, Juan Cao, Peng Qi, Danding Wang, and Jintao Li. 2024.
\newblock \href {https://doi.org/10.1145/3664647.3680663} {{FakingRecipe:
  Detecting Fake News on Short Video Platforms from the Perspective of Creative
  Process}}.
\newblock In \emph{Proceedings of the 32nd ACM International Conference on
  Multimedia}, pages 1351--1360. Association for Computing Machinery.

\bibitem[{Cao et~al.(2020)Cao, Qi, Sheng, Yang, Guo, and Li}]{qipeng-chapter}
Juan Cao, Peng Qi, Qiang Sheng, Tianyun Yang, Junbo Guo, and Jintao Li. 2020.
\newblock \href {https://doi.org/10.1007/978-3-030-42699-6_8} {Exploring the
  Role of Visual Content in Fake News Detection}.
\newblock In \emph{Disinformation, Misinformation, and Fake News in Social
  Media: Emerging Research Challenges and Opportunities}, pages 141--161.
  Springer.

\bibitem[{Chen et~al.(2023)Chen, Xiao, Zhang, Luo, Lian, and Liu}]{bge_m3}
Jianlv Chen, Shitao Xiao, Peitian Zhang, Kun Luo, Defu Lian, and Zheng Liu.
  2023.
\newblock \href {https://arxiv.org/abs/2309.07597} {{BGE M3-Embedding:
  Multi-Lingual, Multi-Functionality, Multi-Granularity Text Embeddings Through
  Self-Knowledge Distillation}}.
\newblock \emph{Preprint}, arXiv:2309.07597.

\bibitem[{Choi and Ko(2021)}]{choi2021cikm}
Hyewon Choi and Youngjoong Ko. 2021.
\newblock \href {https://doi.org/10.1145/3459637.3482212} {{Using Topic
  Modeling and Adversarial Neural Networks for Fake News Video Detection}}.
\newblock In \emph{Proceedings of the 30th ACM International Conference on
  Information \& Knowledge Management}, pages 2950--2954. Association for
  Computing Machinery.

\bibitem[{Hendrickx(2023)}]{hendrickx2023newspapers}
Jonathan Hendrickx. 2023.
\newblock \href {https://doi.org/10.1007/978-3-031-43926-1_16} {From Newspapers
  to TikTok: Social Media Journalism as the Fourth Wave of News Production,
  Diffusion and Consumption}.
\newblock In \emph{Blurring Boundaries of Journalism in Digital Media: New
  Actors, Models and Practices}, pages 229--246. Springer.

\bibitem[{Lei et~al.(2024)Lei, Wang, Ma, Huang, and Liu}]{lei2024comprehensive}
Wentao Lei, Jinting Wang, Fengji Ma, Guanjie Huang, and Li~Liu. 2024.
\newblock \href {https://arxiv.org/abs/2407.08428} {A Comprehensive Survey on
  Human Video Generation: Challenges, Methods, and Insights}.
\newblock \emph{Preprint}, arXiv:2407.08428.

\bibitem[{Leppert and Matsa(2024)}]{TikTokNewsConsumption}
Rebecca Leppert and Katerina~Eva Matsa. 2024.
\newblock More Americans – especially young adults – are regularly getting
  news on TikTok.
\newblock
  \url{https://www.pewresearch.org/short-reads/2024/09/17/more-americans-regularly-get-news-on-tiktok-especially-young-adults/}.

\bibitem[{Li et~al.(2022)Li, Xiao, Li, Hu, Yao, and Li}]{li2022cnn}
Xiaojun Li, Xvhao Xiao, Jia Li, Changhua Hu, Junping Yao, and Shaochen Li.
  2022.
\newblock \href {https://doi.org/10.1038/s41598-022-10117-y} {{A CNN-based
  Misleading Video Detection Model}}.
\newblock \emph{Scientific Reports}, 12(1):1--9.

\bibitem[{Liu et~al.(2024)Liu, Liu, Fu, Wen, Tao, Liu, and
  Li}]{liu2024exploring}
Moyang Liu, Yukun Liu, Ruibo Fu, Zhengqi Wen, Jianhua Tao, Xuefei Liu, and
  Guanjun Li. 2024.
\newblock \href {https://doi.org/10.1109/ISCSLP63861.2024.10800162} {Exploring
  the role of audio in multimodal misinformation detection}.
\newblock In \emph{2024 IEEE 14th International Symposium on Chinese Spoken
  Language Processing}, pages 204--208. IEEE.

\bibitem[{Ma et~al.(2024)Ma, Luo, Guo, Zeng, Hao, and
  Zhao}]{ma-etal-2024-event}
Zihan Ma, Minnan Luo, Hao Guo, Zhi Zeng, Yiran Hao, and Xiang Zhao. 2024.
\newblock \href {https://doi.org/10.18653/v1/2024.acl-long.316} {Event-Radar:
  Event-driven Multi-View Learning for Multimodal Fake News Detection}.
\newblock In \emph{Proceedings of the 62nd Annual Meeting of the Association
  for Computational Linguistics (Volume 1: Long Papers)}, pages 5809--5821.
  Association for Computational Linguistics.

\bibitem[{Nan et~al.(2024)Nan, Sheng, Cao, Hu, Wang, and Li}]{genfend}
Qiong Nan, Qiang Sheng, Juan Cao, Beizhe Hu, Danding Wang, and Jintao Li. 2024.
\newblock \href {https://doi.org/10.1145/3627673.3679519} {Let Silence Speak:
  Enhancing Fake News Detection with Generated Comments from Large Language
  Models}.
\newblock In \emph{Proceedings of the 33rd ACM International Conference on
  Information and Knowledge Management}, pages 1732--1742.

\bibitem[{Niu et~al.(2023)Niu, Lu, Zhang, Cai, Griggio, and
  Heuer}]{niu2023building}
Shuo Niu, Zhicong Lu, Amy~X Zhang, Jie Cai, Carla~F Griggio, and Hendrik Heuer.
  2023.
\newblock \href {https://doi.org/10.1145/3544549.3573809} {Building
  Credibility, Trust, and Safety on Video-Sharing Platforms}.
\newblock In \emph{Extended Abstracts of the 2023 CHI Conference on Human
  Factors in Computing Systems}, pages 1--7. Association for Computing
  Machinery.

\bibitem[{Palod et~al.(2019)Palod, Patwari, Bahety, Bagchi, and
  Goyal}]{palod2019vavd}
Priyank Palod, Ayush Patwari, Sudhanshu Bahety, Saurabh Bagchi, and Pawan
  Goyal. 2019.
\newblock \href {https://doi.org/10.1007/978-3-030-15719-7_18} {Misleading
  Metadata Detection on YouTube}.
\newblock In \emph{Advances in Information Retrieval: ECIR 2019}, pages
  140--147.

\bibitem[{Papadopoulou et~al.(2019)Papadopoulou, Zampoglou, Papadopoulos, and
  Kompatsiaris}]{papadopoulou2019corpus}
Olga Papadopoulou, Markos Zampoglou, Symeon Papadopoulos, and Ioannis
  Kompatsiaris. 2019.
\newblock \href {https://doi.org/10.1108/OIR-03-2018-0101} {A corpus of
  debunked and verified user-generated videos}.
\newblock \emph{Online Information Review}, 43(1):72--88.

\bibitem[{Papadopoulou et~al.(2017)Papadopoulou, Zampoglou, Papadopoulos, and
  Kompatsiaris}]{papadopoulou2017}
Olga Papadopoulou, Markos Zampoglou, Symeon Papadopoulos, and Yiannis
  Kompatsiaris. 2017.
\newblock \href {https://doi.org/10.1145/3078897.3080535} {Web Video
  Verification using Contextual Cues}.
\newblock In \emph{Proceedings of the 2nd International Workshop on Multimedia
  Forensics and Security}, pages 6--10. Association for Computing Machinery.

\bibitem[{{PBS NewsHour Student Reporting Labs}(2023)}]{media-lessons}
{PBS NewsHour Student Reporting Labs}. 2023.
\newblock Media literacy lesson: Three types of manipulated video used to
  spread misinformation.
\newblock
  \url{https://www.pbs.org/newshour/classroom/lesson-plans/2023/03/lesson-plan-three-types-of-manipulated-video-used-to-spread-misinformation}.

\bibitem[{Qi et~al.(2023)Qi, Bu, Cao, Ji, Shui, Xiao, Wang, and
  Chua}]{qi2022fakesv}
Peng Qi, Yuyan Bu, Juan Cao, Wei Ji, Ruihao Shui, Junbin Xiao, Danding Wang,
  and Tat-Seng Chua. 2023.
\newblock \href {https://doi.org/10.1609/aaai.v37i12.26689} {{FakeSV}: A
  Multimodal Benchmark with Rich Social Context for Fake News Detection on
  Short Video Platforms}.
\newblock In \emph{Proceedings of the AAAI Conference on Artificial
  Intelligence}, volume~37, pages 14444--14452. AAAI Press.

\bibitem[{{Qwen Team}(2024)}]{qwen1.5}
{Qwen Team}. 2024.
\newblock \href {https://qwenlm.github.io/blog/qwen1.5/} {Introducing Qwen1.5}.

\bibitem[{Shang et~al.(2021)Shang, Kou, Zhang, and Wang}]{shang2021tiktec}
Lanyu Shang, Ziyi Kou, Yang Zhang, and Dong Wang. 2021.
\newblock \href {https://doi.org/10.1109/BigData52589.2021.9671928} {{A
  Multimodal Misinformation Detector for COVID-19 Short Videos on TikTok}}.
\newblock In \emph{2021 IEEE International Conference on Big Data}, pages
  899--908. IEEE.

\bibitem[{Sheng et~al.(2022)Sheng, Cao, Zhang, Li, Wang, and Zhu}]{zoomout}
Qiang Sheng, Juan Cao, Xueyao Zhang, Rundong Li, Danding Wang, and Yongchun
  Zhu. 2022.
\newblock \href {https://doi.org/10.18653/v1/2022.acl-long.311} {{Zoom Out and
  Observe: News Environment Perception for Fake News Detection}}.
\newblock In \emph{Proceedings of the 60th Annual Meeting of the Association
  for Computational Linguistics (Volume 1: Long Papers)}, pages 4543--4556.
  Association for Computational Linguistics.

\bibitem[{SimPilotSky(2019)}]{simulation}
SimPilotSky. 2019.
\newblock Ethiopia Plane Crash, Ethiopia Airlines B737 MAX Crashes After
  Takeoff, Addis Ababa Airport [XP11].
\newblock \url{https://www.youtube.com/watch?v=q5QnJ9OHkBI&t=574s}.

\bibitem[{Soucek and Lokoc(2024)}]{transnetv2}
Tom\'{a}s Soucek and Jakub Lokoc. 2024.
\newblock \href {https://doi.org/10.1145/3664647.3685517} {TransNet V2: An
  Effective Deep Network Architecture for Fast Shot Transition Detection}.
\newblock In \emph{Proceedings of the 32nd ACM International Conference on
  Multimedia}, page 11218–11221. Association for Computing Machinery.

\bibitem[{Tan et~al.(2024)Tan, Li, Wang, Beigi, Jiang, Bhattacharjee, Karami,
  Li, Cheng, and Liu}]{tan2024large}
Zhen Tan, Dawei Li, Song Wang, Alimohammad Beigi, Bohan Jiang, Amrita
  Bhattacharjee, Mansooreh Karami, Jundong Li, Lu~Cheng, and Huan Liu. 2024.
\newblock \href {https://doi.org/10.18653/v1/2024.emnlp-main.54} {Large
  Language Models for Data Annotation and Synthesis: A Survey}.
\newblock In \emph{Proceedings of the 2024 Conference on Empirical Methods in
  Natural Language Processing}, pages 930--957. Association for Computational
  Linguistics.

\bibitem[{Tang et~al.(2002)Tang, Luo, and Roukos}]{tang-etal-2002-active}
Min Tang, Xiaoqiang Luo, and Salim Roukos. 2002.
\newblock \href {https://doi.org/10.3115/1073083.1073105} {Active Learning for
  Statistical Natural Language Parsing}.
\newblock In \emph{Proceedings of the 40th Annual Meeting of the Association
  for Computational Linguistics}, pages 120--127. Association for Computational
  Linguistics.

\bibitem[{van~der Maaten and Hinton(2008)}]{tsne}
Laurens van~der Maaten and Geoffrey Hinton. 2008.
\newblock {Visualizing Data using t-SNE}.
\newblock \emph{Journal of Machine Learning Research}, 9:2579--2605.

\bibitem[{Wan et~al.(2025)Wan, Wu, Luo, Zeng, and Su}]{wan2025truth}
Herun Wan, Jiaying Wu, Minnan Luo, Zhi Zeng, and Zhixiong Su. 2025.
\newblock \href {https://arxiv.org/abs/2506.02350} {Truth over Tricks:
  Measuring and Mitigating Shortcut Learning in Misinformation Detection}.
\newblock \emph{Preprint}, arXiv:2506.02350.

\bibitem[{Wu et~al.(2024)Wu, Lin, Cao, and Lin}]{wu2024interpretable}
Kaixuan Wu, Yanghao Lin, Donglin Cao, and Dazhen Lin. 2024.
\newblock \href {https://aclanthology.org/2024.lrec-main.804/} {Interpretable
  Short Video Rumor Detection Based on Modality Tampering}.
\newblock In \emph{Proceedings of the 2024 Joint International Conference on
  Computational Linguistics, Language Resources and Evaluation (LREC-COLING
  2024)}, pages 9180--9189. ELRA and ICCL.

\bibitem[{Zeng et~al.(2024)Zeng, Luo, Kong, Liu, Guo, Yang, Ma, and
  Zhao}]{zeng2024mitigating}
Zhi Zeng, Minnan Luo, Xiangzheng Kong, Huan Liu, Hao Guo, Hao Yang, Zihan Ma,
  and Xiang Zhao. 2024.
\newblock \href {https://doi.org/10.1145/3664647.3681673} {{Mitigating World
  Biases: A Multimodal Multi-View Debiasing Framework for Fake News Video
  Detection}}.
\newblock In \emph{Proceedings of the 32nd ACM International Conference on
  Multimedia}, pages 6492--6500. Association for Computing Machinery.

\end{thebibliography}

\end{document}